\documentclass{bmvc2k}

\usepackage{microtype}
\usepackage{graphicx}
\usepackage{subfigure}
\usepackage{booktabs} % for professional tables
\usepackage{amsmath,amssymb}
\usepackage{mathdots}
\usepackage{txfonts}

\title{Resembled Generative Adversarial Networks: Two Domains with Similar Attributes}

\addauthor{Duhyeon Bang}{duhyeonbang@yonsei.ac.kr}{1}
\addauthor{Hyunjung Shim}{kateshim@yonsei.ac.kr}{1}

\addinstitution{
 School of Integrated Technology, Yonsei University, South Korea
}

\runninghead{Duhyeon Bang, Hyunjung Shim}{Resembled GAN}

\def\eg{\emph{e.g}\bmvaOneDot}

\def\etal{\emph{et al}\bmvaOneDot}

\begin{document}

\maketitle

\begin{abstract}
We propose a novel algorithm, namely Resembled Generative Adversarial Networks (GAN), that generates two different domain data simultaneously where they resemble each other. Although recent GAN algorithms achieve the great success in learning the cross-domain relationship \cite{ref02,ref03,ref04}, their application is limited to domain transfers, which requires the input image. The first attempt to tackle the data generation of two domains was proposed by CoGAN \cite{ref05}. However, their solution is inherently vulnerable for various levels of domain similarities. Unlike CoGAN, our Resembled GAN implicitly induces two generators to match feature covariance from both domain, thus leading to share semantic attributes. Hence, we effectively handle a wide range of structural and semantic similarities between various two domains. Based on experimental analysis on various datasets, we verify that the proposed algorithm is effective for generating two domains with similar attributes.

\end{abstract}

%-------------------------------------------------------------------------
\section{Introduction}
\label{sec:intro}

Generative adversarial networks (GANs) are capable of producing sharp and realistic images by learning the generative process, instead of explicitly estimating the data distribution with variational bounds or strict model constraints. GANs \cite{ref01} is composed of two networks, discriminator and generator, and they adversarially compete each other to approximate ${\mathit{P}}_{data}$ using ${\mathit{P}}_{model}$: the discriminator distinguishes real samples from fake samples produced by the generator, while the generator aims to create the sample as real as possible so that the discriminator cannot recognize it as the fake sample. The objective function of this adversarial learning process in \cite{ref01} is defined by the following minimax game, 

{\footnotesize
\[ {\mathop{\mathrm{min}}_{G} \ }{\mathop{\mathrm{max}}_{D} } \mathrm{\ {\mathbb{E}}_{x\sim \mathit{P}_{\mathrm{data}}}\left[\log(D(x))\right]} \ +\ {\mathbb{E}}_{z\sim \mathit{P}_{\mathrm{z}}}\left[\log(1-\mathrm{D}\left(\mathrm{G}\left(\mathrm{z}\right)\right)\right]\ ,\]
}
where $\mathbb{E}$ denotes expectation, $\mathrm{x}$ and $\mathrm{z}$ are random variables for data and latent vector, where their probability distributions are  ${\mathit{P}}_{\mathit{data}}$ and ${\mathit{P}}_{\mathit{z}}$, respectively.

Most GAN algorithms learn an unidirectional mapping function from ${\mathit{P}}_z$ to ${\mathit{P}}_{data}$ for a single domain. Unlike those, our algorithm learns two mapping functions for two domains simultaneously; one associates ${\mathit{P}}_z$ to ${\mathit{P}}^{\mathrm{x}}_{data}$, and the other maps the same ${\mathit{P}}_z$ to ${\mathit{P}}^{\mathrm{y}}_{data}$. Throughout this paper, we denote two different domains by $\mathrm{X}$ and $\mathrm{Y}$, respectively. When $\mathit{x}$ and $\mathit{y}$ are samples generated from the same latent $\mathit{z}$, we aim to accomplish two objectives; 1) two data obey their own data distribution, and 2) two data hold shareable characteristics as similar as possible. For example, faces of human and those of cat represent different species, technically different domains. Hence, they have different shapes and structures. However, their posture, hair color, or facial expression can be similar in both domains, thus those attributes can be regarded as shareable characteristics. Then, our research goal is to generate a pair of human face and cat face that faithfully produce their domain characteristics, and at the same time both faces show the similar attributes such as pose, hair style, or facial color. Note that our algorithm is categorized as an unsupervised GAN because we do not rely on correspondences between two different domains. Also, our approach does not condition on additional input data (i.e., unconditional GAN). Thus, our algorithm is categorized into an unsupervised unconditional GAN for generating two domain data simultaneously.  

Recent studies learn the relationship between two different data domain for domain transfer. They include cycleGAN \cite{ref02}, DiscoGAN \cite{ref03}, and dualGAN \cite{ref04}. Because they aim to establish an image-to-image translation, they require to have the input image as the given condition so to generate the output image in different domain. This problem is inherently analogous to the problem of conditional GAN; it is a different problem from unconditinoal GAN where the data is generated from ${\mathit{P}}_z$. CoGAN \cite{ref05} made the first attempt toward the unsupervised unconditional GAN for generating two domains, as same as our study. They formulate this problem by learning the joint distribution ${\mathit{P}^{x,y}_{data}}$. Suppose the joint probability distribution is factorized as  ${\mathit{P}^{x,y}_{data} = \mathit{P}^x_1 \cdot \mathit{P}^y_2 \cdot \mathit{P}^{x,y}_3}$. CoGAN assumes that $\mathit{P}^{x,y}_3$ is related to high-level semantics while $\mathit{P}^{x}_1$ and $\mathit{P}^{y}_2$ are related to low-level details. Based on this assumption, they suggests a new GAN architecture of two generators. To model $\mathit{P}^{x,y}_3$, first several layers of two generators are coupled by a weight-sharing constraint. The last remaining layers for both generators are designed to learn $\mathit{P}^x_1$ and $ \mathit{P}^y_2$, respectively. It is because the first layers decode high-level semantics and the last layers decode low-level details in the generator. Although each generator is trained with samples for a single data domain, two generators are enforced to share high-level representations during training because of the shared layers. This weight-sharing constraint works well when there is the high structural similarity between two domains. However, with the low structural similarity, the network constraint for CoGAN is too restrictive to achieve the factorization; two generated samples may not show similar attributes.  

Unlike CoGAN, the proposed GAN, namely Resembled GAN, does not explicitly design the network architecture to enforce structural similarity. Instead, our approach employs the feature statistics as an additional constraint, thus it is naturally more flexible to handle a wide range of structural similarities between two data domains.

The main contribution of this study is to define a new objective function of discriminators that leads generators to model the joint distribution, $\mathit{P}^{x,y}_{data}$. To factorize this joint distribution, we propose a feature statistic matching algorithm. Suppose that we derive a feature space where all samples are representative. On this feature space, we assume that the feature distribution of all training data for each domain forms a multivariate Gaussian distribution. After that, we regard a mean vector of each Gaussian as the independent component (i.e. domain specific characteristics), associated with $\mathit{P}^{x}_{1}$ or $\mathit{P}^{y}_{2}$. On the other hand, the covariance matrix represents the dependent component (i.e. shareable attributes), associated with $\mathit{P}^{x,y}_{3}$. Under this assumption, we enforce that two feature distributions have similar feature covariance matrices, effectively leveraging the covariance of two feature distributions.

Using our algorithm, different levels of structural similarity is accounted by the feature covariance; lower the similarity, greater the difference of covariance matrices. As the results, even two domain data are structurally quite different, we can maintain the quality of data generation as well as the attribute similarity.

%-------------------------------------------------------------------------
\section{Background}
%-------------------------------------------------------------------------

We categorize recent studies handling multiple data domains using GANs based on 1) supervised versus unsupervised approach, and 2) conditional versus unconditional approach. 

\textbf{Supervised Conditional GAN (\eg, Image-to-Image translation):} Pix2Pix \cite{ref11} and BicycleGAN \cite{ref12} propose Image-to-Image translation techniques, which transforms the image of the input domain to the image of the target domain. To learn such transformation, they utilize a set of paired images as training data, and they employ the mapping constraint; the transformed image should match the paired image. Finally, they combine a GAN loss with a traditional loss (\eg, $L_1$ or $L_2$) and learn a transformation function that maps X to Y domain. Moreover, they utilize the input image as priors for training the generator and discriminator. The main difference between two studies is a complexity of mapping functions; pix2pix aims to learn a one-to-one mapping while BicycleGAN learns a one-to-many mapping.

\textbf{Unsupervised Conditional GAN (\eg, Learning domain transfer):} CycleGAN \cite{ref02}, DiscoGAN \cite{ref03}, and DualGAN \cite{ref04} are domain transfer algorithms using unpaired two domain images. Because there are no correspondences between images from two domains, they develop two mapping functions (i.e., forward and inverse mapping) and utilize the cycle consistency loss; the sequential operation of forward and inverse mapping should result in the identity mapping. They produce plausible results in learning cross domain relationship. However, they are incapable of generating other domain data without the input. 
% their success also relies on the input image similar to conditional GANs; their problem is different from the original GAN which learns to generate two domain images from a random vector $z \sim {\mathit{P}}_z$.

\textbf{Unsupervised Unconditional GAN (\eg, Learning joint distribution):}
CoGAN \cite{ref05} first suggest the unsupervised unconditional GAN for generating two domain data from ${\mathit{P}}_z$ at the same time, using unpaired two domain data. This is achieved by enforcing a weight-sharing constraint that restricts the generator capacity and favors a joint distribution solution over a product of marginal distributions. Later, the idea of weight sharing was extended to multiple domain translations by Lucic \etal \cite{ref14}. However, this study is categorized as unsupervised conditional GAN because it requires input images for generating corresponding translated images.

%Resembled GAN is unsupervised unconditional GANs for learning joint distribution. However, we do not utilize weight-sharing constraints. Because weight-sharing constraint has clear limitation that structural characteristics between two domain data should be quite similar, as we said in Section ~\ref{sec:intro}. To overcome the limitation, we propose new approach that enforce feature covariance matching on feature space and could extend the range of structural differences between two data domains.

%-------------------------------------------------------------------------
\section{Unsupervised unconditional Resembled GAN}
%-------------------------------------------------------------------------
Our goal is to generate paired images corresponding to two different domains with unpaired training sets. Given two different domain distributions, we aim to train two generators that can faithfully reproduce original characteristics of each domain data distribution (i.e. statistically independent component), and at the same time, retain shareable characteristics (i.e. dependent component of joint distribution) as similar as possible. For that, we define the feature space that represents both domain characteristics using an encoder of a pre-trained autoencoder (AE), and then match two covariance matrices of two feature distributions derived through the encoder.

MGGAN \cite{ref10} first introduced the idea of inducing the generator to possess specific manifold characteristics using a guidance network. Their guidance network leads the generator to learn Forward KL divergence by matching the feature distribution of fake images to that of real images. In this way, they effectively solve a mode collapse problem without sacrificing the image quality of baseline GAN. Inspired by this success, Resembled GAN employs the idea of guidance networks into our problem. That is, we construct a new GAN architecture that two generators produce samples that represent their own domain characteristics and shareable attributes of each other.

\begin{figure}[t!]
  \centering
    \includegraphics[width=0.5\columnwidth]{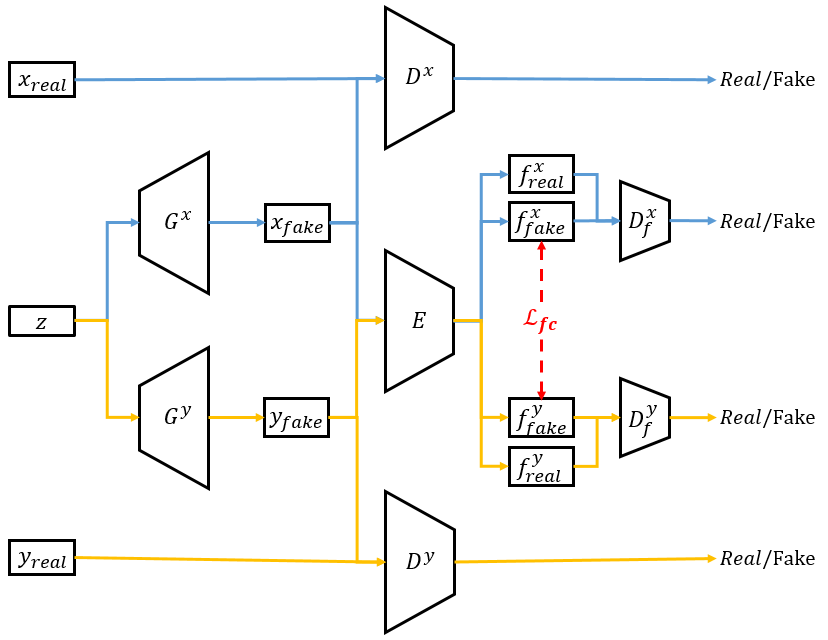}
    % \vskip 0.2in
    \caption{Architecture of Resembled GAN. $\mathrm{X}$ and $\mathrm{Y}$ represent two domain data. $\mathrm{x}_{real}$ ($\mathrm{y}_{real}$) and $\mathrm{x}_{fake}$ ($\mathrm{y}_{fake}$) are the sample of ${\mathit{P}}_{data}$ and ${\mathit{P}}_{model}$, respectively. $\mathrm{z}$ is latent vector. $\mathrm{E}$, $\mathrm{G}$, and  $\mathrm{D}$ are the encoder, the generator, and the discriminator network. The superscript of ${\mathrm{G}}^{\mathrm{x}}$ (${\mathrm{G}}^{\mathrm{y}}$) and ${\mathrm{D}}^{\mathrm{x}}$ (${\mathrm{D}}^{\mathrm{y}}$) indicates the domain and the subscript of ${\mathrm{D}}^{\mathrm{x}}_{f}\ $(${\mathrm{D}}^{\mathrm{y}}_{f}$) indicates the feature space.}
    \label{figure01}
  \vskip -0.2in
\end{figure}

We first define two discriminators for each generator; ${\mathrm{D}}^{\mathrm{x}}$ (${\mathrm{D}}^{\mathrm{y}}$) distinguishes the real distribution from the fake distribution while ${\mathrm{D}}^{\mathrm{x}}_{f}$ (${\mathrm{D}}^{\mathrm{y}}_{f}$) distinguishes the real feature distribution from the fake feature distribution. The feature space is defined by an encoder, $\mathrm{E}$. Then, we introduce a new loss term ${\mathcal{L}}_{\mathrm{fc}}$ that represents the difference between the feature covariance matrix of $\mathrm{x}$ and that of $\mathrm{y}$. We refer this term as the feature covariance constraint, which enables to implicitly learn the joint distribution. Figure~\ref{figure01} visualizes the network architecture for our Resembled GAN. Its overall objective function is summarized as follow. Firstly, the objective of four discriminators (i.e., ${\mathrm{D}}^{\mathrm{x}}$, ${\mathrm{D}}^{\mathrm{y}}$, ${\mathrm{D}}^{\mathrm{x}}_f$, and ${\mathrm{D}}^{\mathrm{y}}_f$) is 

{\footnotesize
\begin{align*}
    \begin{split}
    \mathop{\mathrm{min}}_{{\mathrm{D}}^{\mathrm{x}}, \ {\mathrm{D}}^{\mathrm{y}}, \ {\mathrm{D}}^{\mathrm{x}}_{f},\ {\mathrm{D}}^{\mathrm{y}}_{f}}
    &  \mathop{\mathbb{E}}_{x\sim P^x_{data}}
    \Big[
       \mathrm{log}{\mathrm{D}}^{\mathrm{x}}\left(x\right)
    +  \mathrm{log}{\mathrm{D}}^{\mathrm{x}}_{f}\left(\mathrm{E}\left(x\right)\right)
    \Big] 
    +  \mathop{\mathbb{E}}_{y\sim P^y_{data}}
    \Big[
       \mathrm{log}{\mathrm{D}}^{\mathrm{y}}\left(y\right)
    +  \mathrm{log}{\mathrm{D}}^{\mathrm{y}}_{f}\left(\mathrm{E}\left(y\right)\right)
    \Big] \\
    &+\mathop{\mathbb{E}}_{z\sim P_z} 
    \Big[
       {\mathrm{log} \left(1-{\mathrm{D}}^{\mathrm{x}}\left(\mathrm{G}^x\left(z\right)\right)\right) } 
    +  {\mathrm{log} \left(1-{\mathrm{D}}^{\mathrm{y}}\left(\mathrm{G}^y\left(z\right)\right)\right) } 
    + {\mathrm{log} \left(1-{\mathrm{D}}^{\mathrm{x}}_{f}\left(\mathrm{E}\left(\mathrm{G}^x\left(z\right)\right)\right)\right) } 
    +  {\mathrm{log}\left(1-{\mathrm{D}}^{\mathrm{y}}_{f}\left(\mathrm{E}\left(\mathrm{G}^y\left(z\right)\right)\right)\right) }
    \Big].
    \end{split}
\end{align*}
}
To update the parameters of two generators (i.e., ${\mathrm{G}}^{\mathrm{x}}$ and ${\mathrm{G}}^{\mathrm{y}}$), we optimize the following objective function.
{\footnotesize
\begin{align*}
    \begin{split}
    &\mathop{\mathrm{min}}_{\mathrm{G}^x,\ \mathrm{G}^y} \ \ 
     -\mathop{\mathbb{E}}_{z\sim P_z}
    \Big[
       {\mathrm{log} \left({\mathrm{D}}^{\mathrm{x}}\left(\mathrm{G}^x\left(z\right)\right)\right)\ }  
    +  {\mathrm{log} \left({\mathrm{D}}^{\mathrm{y}}\left(\mathrm{G}^y\left(z\right)\right)\right)\ } 
    + {\mathrm{log} \left({\mathrm{D}}^{\mathrm{x}}_{f}\left(\mathrm{E}\left(\mathrm{G}^x\left(z\right)\right)\right)\right)}
    +  {\mathrm{log} \left({\mathrm{D}}^{\mathrm{y}}_{f}\left(\mathrm{E}\left(\mathrm{G}^y\left(z\right)\right)\right)\right)\ }
    \Big]\ 
    +  \omega {\mathcal{L}}_{fc} , \\
    & \ \ {\mathcal{L}}_{fc} \ \ \ = \ \
    \Big\lVert \ 
    {
      {\mathrm{E}} \left(\mathrm{G}^x\left(z\right)\right)
    - {\mathbb{E}}_{x\sim \mathit{P}^x_{data}}\left[\mathrm{E}\left(x\right)\right] \ 
    ,  \ {\mathrm{E}} \left(\mathrm{G}^y\left(z\right)\right)
    - {\mathbb{E}}_{y\sim P^y_{data}}\left[\mathrm{E}\left(y\right)\right]
    } \
    \Big\lVert_{1}
    \end{split}
\end{align*}
}
\noindent where ${\omega}$ serves the weighting factor for ${\mathcal{L}}_{fc}$. We update the discriminators and the generators alternatively, and two discriminators for each domain, $\mathrm{\ }{\mathrm{D}}^{\mathrm{x}}$ (${\mathrm{D}}^{\mathrm{y}}$) and $\mathrm{\ }{\mathrm{D}}^{\mathrm{x}}_{f}$ (${\mathrm{D}}^{\mathrm{y}}_{f}$), are trained independently. %We explain the details of the encoder and this constraint in subsection.

\subsection{Analyzing the feature distribution}
To utilize the statistics of feature distribution as constraints, we first develop the feature space where all samples from two domains are analyzed. To compare feature statistics from two domains, we should ensure that this feature space should be representative for all samples from both domains. Suppose that they represent only the major mode in each domain, or are biased toward one of two domains in feature space. In such a case, generators suffer from mode collapse or fail to learn the other domain characteristic. To cope all data from both domains, we utilize an encoder from a pre-trained AE. Because AE aims to reconstruct all training samples, the encoded features from AE faithfully represents the data distribution \cite{ref15}. Using all samples from both domains as training set, we can ensure that the AE can encode all samples into the same feature space; two feature distributions from both domains can be comparable. Furthermore, we modify the AE to the denoising AE in order to improve the robustness of model as a feature extractor \cite{ref16}. Finally, we pre-train and fix the parameters of the AE during GAN training. In this way, we maintain the representative power of feature space defined by its encoder. 
%  \vspace{-3mm}
\subsection{Feature covariance constraint, ${\mathcal{L}}_{fc}$ }

Our key idea is to enforce the feature covariance constraint for GAN training, implicitly learning the joint distribution. The similar idea has been discussed in the previous study, Snell \etal \cite{ref17} for few-shot learning. Assuming the feature distribution on embedding space as a multivariate Gaussian distribution, they claim that the mean of real feature distribution represents the identity attribute of the class while the covariance of that represents intra-class variation (\eg, shareable attributes). Inspired by this observation, we regard the mean of feature distribution as domain identity attribute, and the covariance of that as shareable attributes. From the similar observation, several studies for low-shot learning augment the data of long-tailed classes by referencing the feature covariance of rich class data. Given a single or few training image, they manipulate their features of training set to generate additional training data. For example, Yin \etal \cite{ref18} transfer the principal components from regular classes to tail classes so to increase their intra-classes variance.

% Since ${\mathrm{D}^{\mathrm{y}}_{f}}$ (${\mathrm{D}^{\mathrm{y}}_{f}}$) reduces divergence between feature distribution of real and that of fake, we could define a feature mean vector, i.e. class identity vector, through expectation of real feature vector, ${ {\mathbb{E}}_{x\sim \mathit{P}^x_{data}}[ \mathrm{E}\left(x\right)] }$ ( ${ {\mathbb{E}}_{y\sim \mathit{P}^y_{data}}[ \mathrm{E}\left(y\right)] }$ ). As the results, feature covariance vectors are defined by subtracting the mean feature vector from fake feature vector, ${{\mathrm{E}} \left(\mathrm{G}^x\left(z\right)\right)}$ (${  {\mathrm{E}} \left(\mathrm{G}^y\left(z\right)\right) }$), where ${z \sim {\mathit{P}}_z}$. Finally, generators learn joint distribution by minimizing difference between feature covariance. 
% \vspace{-3mm}
%-------------------------------------------------------------------------
\section{Evaluation}
%-------------------------------------------------------------------------
In this section, we first analyze the performance of the Resembled GAN by 1) adapting various domains, 2) conducting the image reconstruction, and 3) generating by the latent space walking. Then, we compare our model with CoGAN, which is a baseline algorithm and evaluate how well their generation retains the semantic similarity between two domains. Also, we quantitatively evaluate the generation quality both in terms of diversity and image quality. 

%For qualitative and quantitative evaluation, we use CoGAN as a baseline algorithm and compare it with Resembled GAN. This is because CoGAN is the only unsupervised unconditional GANs for generating two domain data simultaneously, as same as ours. 
%CoGAN accomplishes this goal by enforcing weight-sharing of several generator layers that decode high-level semantics. However, their weight-sharing constraint also requires the strict condition that two domains should present the high structural similarity. Resembled GAN overcomes this limitation with a feature covariance constraint.

To confirm whether each model can handle a wide range of structural similarities across domain, we experiment with two scenarios.  
 \vspace{-1mm}
\begin{enumerate}
\item High structural and semantical similarity: We divide CelebA \cite{ref19} dataset into two domains using its attribute labels; such as gender, hair colors or with/without glasses. This scenario is relatively easy because any pair of domains have high structural similarity (i.e., all are human faces) with the precise alignment.
 \vspace{-1mm}
\item Low structural and semantical similarity: We choose the CelebA dataset and Cat head dataset \cite{ref20} to construct the problem of handling significantly different two domains. This scenario is relatively hard because human and cat face has low structural similarity with the poor alignment. 
\end{enumerate}
For fair comparison, we design the architecture of discriminator and that of generator for Resembled GAN and CoGAN based on DCGAN \cite{ref22}; for that, we crop and resize all dataset into $64\times64\times3$.
The cat head dataset was extracted by using provided facial key points in a 10k cat images; 9k images for training and 1k images for testing. For improving the training stability, we increase the training dataset to 90k using the affine transform based data augmentation. We randomly draw 90k images from the CelebA dataset to match the number of Cat head data. 
To implement CoGAN, except the last layer of generator and the first layer of discriminator, the rest of layers from each domain network is tied for a weight-sharing constraint. The architecture of manifold discriminators, ${\mathrm{D}}^{\mathrm{x}}_{f} $ and ${\mathrm{D}}^{\mathrm{y}}_{f}$, follow that of MGGAN. We set $\omega=1$ for all experimental evaluations.  

\subsection{Evaluating the semantic similarity between two domains}

\begin{figure}[t!]
  \centering
    \includegraphics[width=\columnwidth]{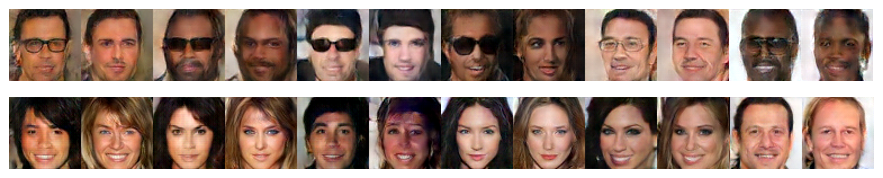}
    % \vskip 0.2in
    \caption{Generation with high structural and semantical similarity dataset; with/without glasses(top), black/blond hair(bottom). Each of odd and even columns are resembled image generated from same random noise vectors.}
    \label{figure02}
  \vskip -0.2in
\end{figure}

First, we evaluate the Resembled GAN with various paired domains: with/without glasses and black/blond hair styles. Our generation results are summarized in Fig.~\ref{figure02}. These generated images present the domain characteristics reasonably well such as the presence of glasses or color of hair while they hold shareable attributes such as the rest of facial structure. 

\begin{figure}[t!]
  % \vskip 0.2in
  \centering
    \includegraphics[width=\columnwidth]{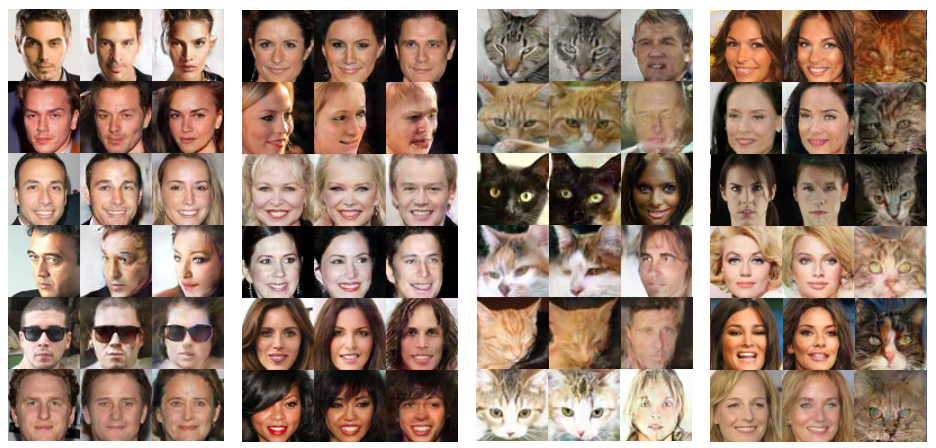}
    \caption{Reconstruction results. Each of the first, second, and third columns are real test image, a reconstructed image, and a resemble image of the corresponding domain.}
    \label{figure03}
  \vskip -0.2in
\end{figure}

Resembled GAN is capable of reconstructing real data by introducing a small modification to the network. The image reconstruction has originally been demonstrated in MGGAN. That is, we add the three layers of fully connected network to map ${\mathit{P}}_f$ to ${\mathit{P}}_z$. To perform the image reconstruction with CoGAN, they should conduct the additional optimization for searching a latent vector corresponding to real data. Unfortunately, this optimization is error prone and unreliable because it should solve inverse generation process, which is extremely non-linear and complicated like the generation process. Using Resembled GAN, the reconstruction can be easily formulated as a part of generative model, thus the reconstruction results are more plausible than optimization based reconstruction. Fig.~\ref{figure03} shows our reconstruction results. Each of the first, second, and third column are real images, reconstructed images and generated images in the other domain. 

\begin{figure}[t!]
%   \vskip 0.2in
  \centering
    \includegraphics[width=\columnwidth]{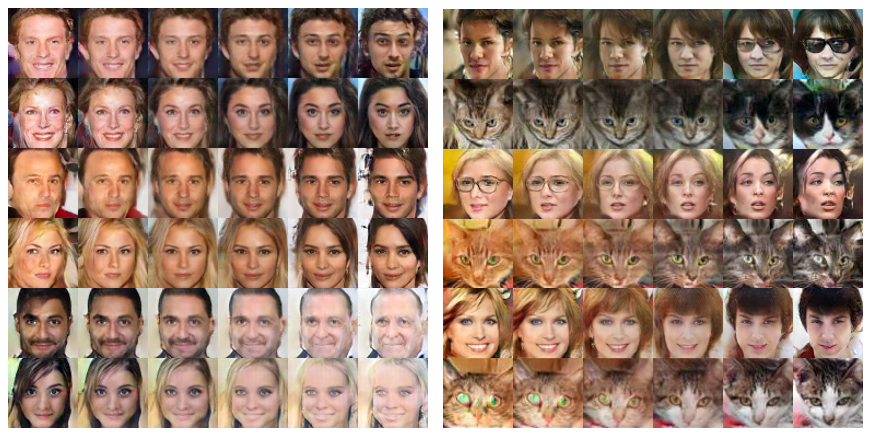}
    % \vskip 0.2in
    \caption{Interpolation results. Each of odd and even rows are pair images generated from same interpolated latent vectors.}
    \label{figure04}
  \vskip -0.2in
\end{figure}

To verify whether data generation is the results of data memorization or not, we generate samples by latent space walking. The generated images from interpolated latent vectors between two specific vectors do not have meaningful connectivity if the generator just memorize the dataset, such as lack of smooth transitions or fail to generation \cite{ref21,ref22}. From semantically smooth interpolation results shown in Fig.~\ref{figure04}, we conclude Resembled GAN reproduces the data distribution without memorization. More interestingly, we observe that latent walking in two domains demonstrates semantically similar. For example, in the middle row of Fig.~\ref{figure04}, the smooth transitions of facial poses are quite similar in both domains. We have the same observation consistently over various examples.

% \vspace{-2mm}
\begin{figure}[t!]
%   \vskip 0.2in
  \centering
    \includegraphics[width=\columnwidth]{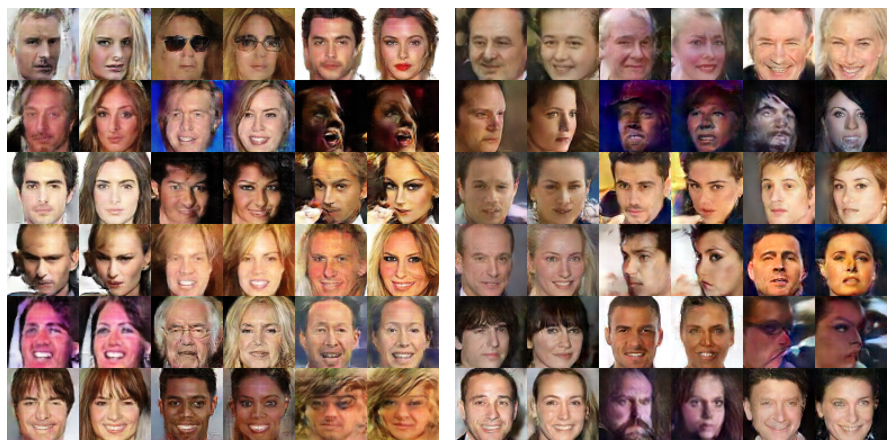}
    \includegraphics[width=\columnwidth]{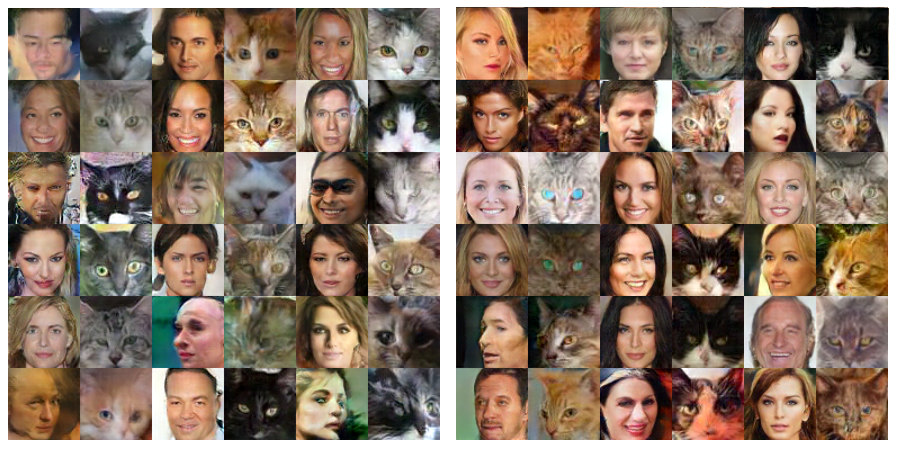}
    % \vskip 0.2in
    \caption{Comparison between CoGAN (Left) and Resembled GAN (Right) using two cases of dataset: one (top) utilize male and female of CelebA; the other (bottom) utilize celebA and cat face dataset.}
    \label{figure05}
  \vskip -0.2in
\end{figure}
% \vspace{-2mm}
Finally, we qualitatively compare Resembled GAN with CoGAN. For fair comparison, we draw samples at the same iteration of training, 35k. The generated images are shown in Fig.~\ref{figure05}; the left half are from CoGAN, and the right half are from Resembled GAN. Each of odd and even column images are paired, each generated from  ${\mathrm{G}}^{\mathrm{x}}$ and ${\mathrm{G}}^{\mathrm{y}}$ from the same random vectors,  ${\mathrm{z} \sim \mathit{P}_z}$. 

When generating the different gender, the quality of CoGAN and Resembled GAN are nearly the same. Both algorithms retain each domain characteristic clearly (male and female) while keeping shareable attributes; such as smile, skin color or background. An interesting observation is that some attributes associated with the domain characteristics, such as mustache, are excluded from shareable features automatically by networks decision. On the contrary, when handling human and cat faces, Resembled GAN clearly is better than CoGAN, especially how well two generated images share common attributes. Because CoGAN generates images with the weight-sharing constraint, several results can also match the face orientation. However, they do not share hair color, facial shape or eye shapes, which are properly modeled shareable attributes in Resembled GAN. The generated pairs from Resembled GAN can match the facial orientation, background color, hair color, skin tone, facial shape or eye shape (e.g., oval or line-shape). %Although it is hard to judge similarity between them exactly, Resembled GAN seem to share those attributes better than CoGAN.

Although CoGAN aims to match high-level semantics, weight-sharing constraint just considers the structural similarity in substance. Thus, their approach is less effective when structural features between two domains are substantially different. Unlike CoGAN, Resembled GAN constrains the feature statistics. Hence, the strength of constraint is automatically determined by the discrepancy of two domains; the bigger the difference, the stronger the constraint. For this reason, our model is more robust to handle various levels of structural and semantic similarities between two domains.

\subsection{Quantitative Evaluation}
Using existing evaluation metric, it is hard to quantitatively evaluate the semantic similarity of two domains. Instead, we utilize two general metrics for evaluating GANs; one is MS-SSIM for measuring image diversity (\eg, the lower the MS-SSIM, the higher the diverse \cite{ref25,ref26}) and the other is a Fr\'echet distance (FID) for measuring visual quality. (\eg, the higher the FID score, the higher the quality \cite{ref23,ref24})

When training the generators of two domains by keeping the shareable attributes, the generation process tends to increase its intra-class variations because it learns the feature covariance from both domains. As a result, the diversity of generated images in each domain becomes higher than that of real data if generators are influenced by the shareable attributes from the other domain. We observe the same phenomenon in our experiment. For example, although faces of cat in training dataset do not possess red hairs, Resembled GAN can generate cat with red hairs as shown in the last row in Fig.~\ref{figure05}.  This observation consistently holds in the quantitative evaluation summarized in Table~\ref{table01}. All scores are the average MS-SSIM repeated five times for each model and dataset. Resembled GAN achieves greater diversity (i.e., lower MS-SSIM) than all real dataset and CoGAN. From these results, we justify that Resemble GAN possesses the representation power for generating the wide range of attributes, more flexible to model various attributes. 

Because the trade-off relationship between visual quality and sample diversity is a well-known issue in GAN training \cite{ref27}, we also verify whether our achievement in image diversity is the result of sacrificing the image quality. The FID scores in Table~\ref{table01} show that Resembled GAN reports a slightly lower FID than CoGAN in average. However, because the one standard deviation of FID from CoGAN overlaps with Resembled GAN, the statistical difference is not significantly meaningful.

\begin{table}[t!]
  \centering
  \resizebox{.8\columnwidth}{!}{
    \begin{tabular}{|c|c|c c|c c|}
        \hline
            Metric  & Model               & Male    & Female  & Human   & Cat     \\
            \hline\hline
                    & Real-dataset        & 0.3558  & 0.4214  & 0.3897  & 0.2134  \\
            MS-SSIM & Coupled \ \ \ \ GAN & 0.3584  & 0.4351  & 0.3961  & 0.2123  \\
            (mean)   & Resembled GAN       & 0.3392  & 0.4090  & 0.3324  & 0.2069  \\
            \hline
            FID score& Coupled \ \ \ \ GAN & 34.55$\pm$2.45 & 29.59$\pm$3.34 & 38.74$\pm$3.32 & 31.45$\pm$4.21 \\
            (mean $\pm$ std)   & Resembled GAN       & 36.35$\pm$2.21 & 37.33$\pm$4.02 & 41.89$\pm$3.21 & 33.89$\pm$4.59 \\
        \hline
    \end{tabular}}
  \vskip 0.2in
  \caption{Comparison of the sample diversity and quality using MS-SSIM (mean) and FID score (mean and standard deviation), respectively. The lower MS-SSIM, the higher diversity. The higher FID score, the higher quality.}
  \label{table01}
\end{table}
\vspace{-5mm}

%-------------------------------------------------------------------------
\section{Conclusion}
%-------------------------------------------------------------------------
This paper introduces Resembled GAN that generates a pair of images from two domains with similar attributes. The objective of our study is different from those of domain transfer techniques in that we deal with unsupervised and unconditional approach to generating two domains simultaneously. While existing method for the same objective, CoGAN, explicitly enforces the structural similarity between two domains, we induce generators to learn the shareable attribute from the other domain based on feature covariance matching. In this way, Resembled GAN handles semantic attributes such as color mood better than CoGAN. More importantly, Resembled GAN is more flexible to handle various levels of similarities between two domains. We expect that our feature matching idea can be extended toward cross-domain transfers in a unsupervised unconditional manner.


\begin{thebibliography}{22}
\providecommand{\natexlab}[1]{#1}
\providecommand{\url}[1]{\texttt{#1}}
\expandafter\ifx\csname urlstyle\endcsname\relax
  \providecommand{\doi}[1]{doi: #1}\else
  \providecommand{\doi}{doi: \begingroup \urlstyle{rm}\Url}\fi

\bibitem[Bang and Shim(2018)]{ref10}
Duhyeon Bang and Hyunjung Shim.
\newblock Mggan: Solving mode collapse using manifold guided training.
\newblock \emph{arXiv preprint arXiv:1804.04391}, 2018.

\bibitem[Dinh et~al.(2016)Dinh, Sohl-Dickstein, and Bengio]{ref21}
Laurent Dinh, Jascha Sohl-Dickstein, and Samy Bengio.
\newblock Density estimation using real nvp.
\newblock \emph{arXiv preprint arXiv:1605.08803}, 2016.

\bibitem[Fedus et~al.(2017{\natexlab{a}})Fedus, Rosca, Lakshminarayanan, Dai,
  Mohamed, and Goodfellow]{ref26}
William Fedus, Mihaela Rosca, Balaji Lakshminarayanan, Andrew~M Dai, Shakir
  Mohamed, and Ian Goodfellow.
\newblock Many paths to equilibrium: Gans do not need to decrease adivergence
  at every step.
\newblock \emph{arXiv preprint arXiv:1710.08446}, 2017{\natexlab{a}}.

\bibitem[Fedus et~al.(2017{\natexlab{b}})Fedus, Rosca, Lakshminarayanan, Dai,
  Mohamed, and Goodfellow]{ref27}
William Fedus, Mihaela Rosca, Balaji Lakshminarayanan, Andrew~M Dai, Shakir
  Mohamed, and Ian Goodfellow.
\newblock Many paths to equilibrium: Gans do not need to decrease adivergence
  at every step.
\newblock \emph{arXiv preprint arXiv:1710.08446}, 2017{\natexlab{b}}.

\bibitem[Goodfellow et~al.(2014)Goodfellow, Pouget-Abadie, Mirza, Xu,
  Warde-Farley, Ozair, Courville, and Bengio]{ref01}
Ian Goodfellow, Jean Pouget-Abadie, Mehdi Mirza, Bing Xu, David Warde-Farley,
  Sherjil Ozair, Aaron Courville, and Yoshua Bengio.
\newblock Generative adversarial nets.
\newblock In \emph{Advances in neural information processing systems}, pages
  2672--2680, 2014.

\bibitem[Heusel et~al.(2017)Heusel, Ramsauer, Unterthiner, Nessler, Klambauer,
  and Hochreiter]{ref23}
Martin Heusel, Hubert Ramsauer, Thomas Unterthiner, Bernhard Nessler,
  G{\"u}nter Klambauer, and Sepp Hochreiter.
\newblock Gans trained by a two time-scale update rule converge to a nash
  equilibrium.
\newblock \emph{arXiv preprint arXiv:1706.08500}, 2017.

\bibitem[Isola et~al.(2017)Isola, Zhu, Zhou, and Efros]{ref11}
Phillip Isola, Jun-Yan Zhu, Tinghui Zhou, and Alexei~A Efros.
\newblock Image-to-image translation with conditional adversarial networks.
\newblock \emph{arXiv preprint}, 2017.

\bibitem[Kim et~al.(2017)Kim, Cha, Kim, Lee, and Kim]{ref03}
Taeksoo Kim, Moonsu Cha, Hyunsoo Kim, Jungkwon Lee, and Jiwon Kim.
\newblock Learning to discover cross-domain relations with generative
  adversarial networks.
\newblock \emph{arXiv preprint arXiv:1703.05192}, 2017.

\bibitem[Liu and Tuzel(2016)]{ref05}
Ming-Yu Liu and Oncel Tuzel.
\newblock Coupled generative adversarial networks.
\newblock In \emph{Advances in neural information processing systems}, pages
  469--477, 2016.

\bibitem[Liu et~al.(2017)Liu, Breuel, and Kautz]{ref14}
Ming-Yu Liu, Thomas Breuel, and Jan Kautz.
\newblock Unsupervised image-to-image translation networks.
\newblock In \emph{Advances in Neural Information Processing Systems}, pages
  700--708, 2017.

\bibitem[Liu et~al.(2015)Liu, Luo, Wang, and Tang]{ref19}
Ziwei Liu, Ping Luo, Xiaogang Wang, and Xiaoou Tang.
\newblock Deep learning face attributes in the wild.
\newblock In \emph{Proceedings of the IEEE International Conference on Computer
  Vision}, pages 3730--3738, 2015.

\bibitem[Lucic et~al.(2017)Lucic, Kurach, Michalski, Gelly, and
  Bousquet]{ref24}
Mario Lucic, Karol Kurach, Marcin Michalski, Sylvain Gelly, and Olivier
  Bousquet.
\newblock Are gans created equal? a large-scale study.
\newblock \emph{arXiv preprint arXiv:1711.10337}, 2017.

\bibitem[Odena et~al.(2016)Odena, Olah, and Shlens]{ref25}
Augustus Odena, Christopher Olah, and Jonathon Shlens.
\newblock Conditional image synthesis with auxiliary classifier gans.
\newblock \emph{arXiv preprint arXiv:1610.09585}, 2016.

\bibitem[Radford et~al.(2015)Radford, Metz, and Chintala]{ref22}
Alec Radford, Luke Metz, and Soumith Chintala.
\newblock Unsupervised representation learning with deep convolutional
  generative adversarial networks.
\newblock \emph{arXiv preprint arXiv:1511.06434}, 2015.

\bibitem[Rosca et~al.(2017)Rosca, Lakshminarayanan, Warde-Farley, and
  Mohamed]{ref15}
Mihaela Rosca, Balaji Lakshminarayanan, David Warde-Farley, and Shakir Mohamed.
\newblock Variational approaches for auto-encoding generative adversarial
  networks.
\newblock \emph{arXiv preprint arXiv:1706.04987}, 2017.

\bibitem[Snell et~al.(2017)Snell, Swersky, and Zemel]{ref17}
Jake Snell, Kevin Swersky, and Richard Zemel.
\newblock Prototypical networks for few-shot learning.
\newblock In \emph{Advances in Neural Information Processing Systems}, pages
  4080--4090, 2017.

\bibitem[Vincent et~al.(2010)Vincent, Larochelle, Lajoie, Bengio, and
  Manzagol]{ref16}
Pascal Vincent, Hugo Larochelle, Isabelle Lajoie, Yoshua Bengio, and
  Pierre-Antoine Manzagol.
\newblock Stacked denoising autoencoders: Learning useful representations in a
  deep network with a local denoising criterion.
\newblock \emph{Journal of Machine Learning Research}, 11\penalty0
  (Dec):\penalty0 3371--3408, 2010.

\bibitem[Yi et~al.(2017)Yi, Zhang, Tan, and Gong]{ref04}
Zili Yi, Hao Zhang, Ping Tan, and Minglun Gong.
\newblock Dualgan: Unsupervised dual learning for image-to-image translation.
\newblock \emph{arXiv preprint}, 2017.

\bibitem[Yin et~al.(2018)Yin, Yu, Sohn, Liu, and Chandraker]{ref18}
Xi~Yin, Xiang Yu, Kihyuk Sohn, Xiaoming Liu, and Manmohan Chandraker.
\newblock Feature transfer learning for deep face recognition with long-tail
  data.
\newblock \emph{arXiv preprint arXiv:1803.09014}, 2018.

\bibitem[Zhang et~al.(2008)Zhang, Sun, and Tang]{ref20}
Weiwei Zhang, Jian Sun, and Xiaoou Tang.
\newblock Cat head detection-how to effectively exploit shape and texture
  features.
\newblock In \emph{European Conference on Computer Vision}, pages 802--816.
  Springer, 2008.

\bibitem[Zhu et~al.(2017{\natexlab{a}})Zhu, Park, Isola, and Efros]{ref02}
Jun-Yan Zhu, Taesung Park, Phillip Isola, and Alexei~A Efros.
\newblock Unpaired image-to-image translation using cycle-consistent
  adversarial networks.
\newblock \emph{arXiv preprint arXiv:1703.10593}, 2017{\natexlab{a}}.

\bibitem[Zhu et~al.(2017{\natexlab{b}})Zhu, Zhang, Pathak, Darrell, Efros,
  Wang, and Shechtman]{ref12}
Jun-Yan Zhu, Richard Zhang, Deepak Pathak, Trevor Darrell, Alexei~A Efros,
  Oliver Wang, and Eli Shechtman.
\newblock Toward multimodal image-to-image translation.
\newblock In \emph{Advances in Neural Information Processing Systems}, pages
  465--476, 2017{\natexlab{b}}.

\end{thebibliography}
\end{document}